\documentclass[preprint,12pt]{elsarticle}

\usepackage{amssymb}
\usepackage{amsmath}
\usepackage{booktabs}
\usepackage{siunitx}
\usepackage{caption}
\usepackage{placeins}
\usepackage{tabularx}
\usepackage{ragged2e}
\usepackage{amsfonts}
\usepackage{algorithm}
\usepackage{algpseudocode}

\usepackage{graphicx} 
\usepackage{rotating} 
\usepackage{booktabs} 
\usepackage{subcaption}    

\journal{}

\begin{document}

\begin{frontmatter}



\title{FlightKooba: A Fast Interpretable FTP Model} 

\author[1,2]{Jing Lu \corref{cor1}} 
\cortext[cor1]{Corresponding author}
\ead{lujing_cafuc@nuaa.edu.cn}

\author[1]{Xuan Wu}
\author[1]{Yizhun Tian}
\author[1]{Songhan Fan}
\author[1]{Yali Fang}

\affiliation[1]{organization={School of Computer Science and Artificial Intelligence},
            addressline={Civil Aviation Flight University of China}, 
            city={Guanghan},
            postcode={618000}, 
            state={Sichuan},
            country={China}}
\affiliation[2]{organization={College of Computer Science},
            addressline={Technology Nanjing University of Aeronautics and Astronautics Nanjing}, 
            city={Nanjing},
            postcode={210000}, 
            state={Jiangsu},
            country={China}}
            
\begin{abstract}
Flight trajectory prediction (FTP) and similar time series tasks typically require capturing smooth latent dynamics hidden within noisy signals. However, existing deep learning models face significant challenges of high computational cost and insufficient interpretability due to their complex black-box nature. This paper introduces FlightKooba, a novel modeling approach designed to extract such underlying dynamics analytically. Our framework uniquely integrates HiPPO theory, Koopman operator theory, and control theory. By leveraging Legendre polynomial bases, it constructs Koopman operators analytically, thereby avoiding large-scale parameter training. The method's core strengths lie in its exceptional computational efficiency and inherent interpretability. Experiments on multiple public datasets validate our design philosophy: for signals exhibiting strong periodicity or clear physical laws (e.g., in aviation, meteorology, and traffic flow), FlightKooba delivers competitive prediction accuracy while reducing trainable parameters by several orders of magnitude and achieving the fastest training speed. Furthermore, we analyze the model's theoretical boundaries, clarifying its inherent low-pass filtering characteristics that render it unsuitable for sequences dominated by high-frequency noise. In summary, FlightKooba offers a powerful, efficient, and interpretable new alternative for time series analysis, particularly in resource-constrained environments.
\end{abstract}

\begin{highlights}
\item A novel modeling method is proposed, achieving reduced time and memory overhead
\item The method exhibits strong interpretability
\item The model demonstrates competitive performance on datasets exhibiting clear periodicity or physical regularities.
\item The study provides a new option for time series prediction
\end{highlights}


\end{frontmatter}




\section{Introduction}

Flight trajectory prediction (FTP) is one of the fundamental research tasks in the field of aviation \cite{ftp}. It is a key technology for various applications in the fields of flight operation quality assurance (FOQA) \cite{foqa} and air traffic control (ATC) \cite{atc}. Flight trajectory prediction has always been an important aspect of air traffic management and safety. In flight training, trajectory prediction can be used to identify potential risks (such as attitude loss of control and altitude deviation) in advance by analyzing the trainee's operating habits and the aircraft's status (such as rudder deflection angle and roll rate). Flight trajectory prediction is not only a technical tool in training, but also a key factor in reshaping pilot training models. By integrating deep learning methods with physical models and combining real-time data-driven optimization strategies, we achieve comprehensive capability enhancements across the entire chain, driving the development of aviation training toward intelligence and precision. Currently, many scholars have introduced deep learning and neural networks to study flight trajectory prediction. However, existing deep learning-based FTP models still face challenges related to computational cost and memory overhead, especially when dealing with long-term temporal dependencies.

Deep learning methods have achieved relatively good success in modeling large complex systems using the \cite{wang2024} dataset. Existing deep learning techniques can be applied across various domains to forecast time series data with excellent results, such as weather forecasting \cite{verma2404} and traffic prediction \cite{li2023}. Koopman theory \cite{koopman1931} has garnered increasing attention in recent years, emerging as a promising data-driven approach. By representing nonlinear systems using linear state-space models (SSM) \cite{kalman1960}, the Koopman operator transforms complex nonlinear dynamic systems into linear forms. Building upon this approach, Williams proposed an Extended Dynamic Modal Decomposition (EDMD) method to compute the Koopman operator \cite{williams2015(edmd)}. With the emergence and advancement of deep learning, Koopman theory has been combined with neural networks across numerous domains, achieving significant success in modeling many complex systems, such as probabilistic neural networks \cite{han2022desko} and graph neural networks \cite{li2019}. {li2019}.

Existing temporal data processing models (e.g., xLSTM, Mamba, etc.) or various models combined with Koopman algorithms have made tremendous progress in their respective domains. However, they all suffer from one or more of the following issues: Computational Intensity: High parameter count and slow training; Black-box nature: Difficult to interpret and trust, posing a critical flaw in safety-critical domains like aviation; Overfitting noise: These powerful models tend to learn every minor fluctuation in the data, encompassing both genuine dynamics and substantial meaningless sensor noise. In many physical systems, we are more concerned with the intrinsic core dynamics of the system than with the noise itself.

To address these issues, we propose a fundamentally different modeling approach. We contend that an excellent dynamical model should not aim to blindly fit every data point, but rather capture and predict the core patterns of system evolution. Inspired by this, we developed FlightKooba, a lightweight framework grounded in classical dynamical systems theory. It maps time series onto a set of Legendre polynomial coefficients via the HIPPO theory, and linearizes complex nonlinear dynamics using Koopman operator theory. Crucially, the system's state transition matrix (i.e., the Koopman operator) is analytically derived from these coefficients, rather than being obtained through brute-force fitting via backpropagation. This design yields two major advantages: first, exceptionally high computational efficiency; second, clear physical and mathematical interpretability.

The main contributions of this work are as follows:

\begin{itemize}
\item A Novel Modeling Approach: We introduce FlightKooba, a novel method for time series forecasting that shifts focus from black-box fitting to the analytical construction of dynamic operators. By integrating HiPPO, Koopman, and control theory, we provide a theoretically grounded approach for modeling underlying system dynamics.
\item Significant Gains in Computational Efficiency: We demonstrate that FlightKooba achieves a reduction in parameter count by several orders of magnitude compared to baseline models like Mamba and xLSTM. This directly translates to substantially faster training times and minimal memory consumption, making it ideal for resource-constrained applications.
\item Competitiveness on target datasets: Experiments demonstrate that for time series with strong periodic patterns or clear physical foundations, FlightKooba achieves an outstanding trade-off between performance and efficiency. It delivers competitive prediction accuracy while substantially reducing computational costs.
\item In-depth analysis of applicability boundaries: We conduct a rigorous analysis of the model's theoretical limitations. We identify and explain its inherent low-pass filtering characteristics, providing clear guidance on its ideal application scenarios and elucidating why it is unsuitable for high-frequency noise-driven signals.
\end{itemize}

The remainder of this paper is organized as follows: Section 2 reviews related work. Section 3 introduces the FlightKooba framework in detail. Section 4 presents the experimental setup, results, and a dedicated discussion on the model's limitations. Finally, Section 5 concludes the paper and outlines future research directions.

\section{Related Works}

Since its introduction, the Koopman operator theory \cite{koopman1931} has been considered a powerful tool for analyzing dynamical systems. In the paper “Identification of MIMO Wiener-type Koopman models for data-driven model reduction using deep learning” \cite{schulze2022}, the scope of Koopman operator theory was extended to controlled systems, and some achievements were made in both modeling and control \cite{zhang2023}. Similarly, when combined with neural network models, this new model with Koopman operators has achieved effective results in various fields such as robotics \cite{shi2022}, fluid physics \cite{morton2018}, and vehicle systems \cite{chen2024}.

Although the Koopman model has received widespread attention, the current methods for solving Koopman operators can be broadly categorized into dynamic mode decomposition (DMD) \cite{schmid2010dmd} and its extensions (EDMD) \cite{williams2015edmd}, Hankel dynamic mode decomposition \cite{arbabi2017hankeldmd}, spectral analysis, and computational methods using machine learning algorithms. For the first three data-driven methods, some computation is required to construct an approximation of the Koopman operator, which incurs additional time overhead. In contrast, our method can approximate the Koopman operator using polynomial approximation, thereby greatly reducing the time required to calculate the Koopman operator.

With the development of data-driven methods, models combining Koopman operators with neural networks have improved modeling accuracy and promoted the application of this model in various fields. In particular, in the field of learning control, various algorithms have been proposed to train corresponding control strategies in established models \cite{ha2018, hafner2019, hafner2020, hafner2023}. There is also a new modeling method (MamKO) that combines the recently popular Mamba model with the Koopman operator \cite{mamko}.

We analyzed and learned from the history of the Mamba structure and discovered its predecessor: the HIPPO algorithm used in the SSM module of the structured state space sequence model (S4) \cite{gu2021efficiently}. It gives the SSM module a physical meaning: use a polynomial approximation of a time function and calculate the coefficient in front of each polynomial. Inspired by this method, we found that for the polynomial approximation equation of a certain function, we can use the idea of deriving state space equations from state equations in “Principles of Automatic Control” to obtain the corresponding form of the Koopman operator. This is equivalent to directly deriving the approximate form of the Koopman operator from polynomial approximation, which greatly reduces the parameters of neural network calculations while ensuring a certain degree of accuracy.

\section{Methodology}

\begin{figure}[!hbt]
\centering
\includegraphics[width=0.8\linewidth]{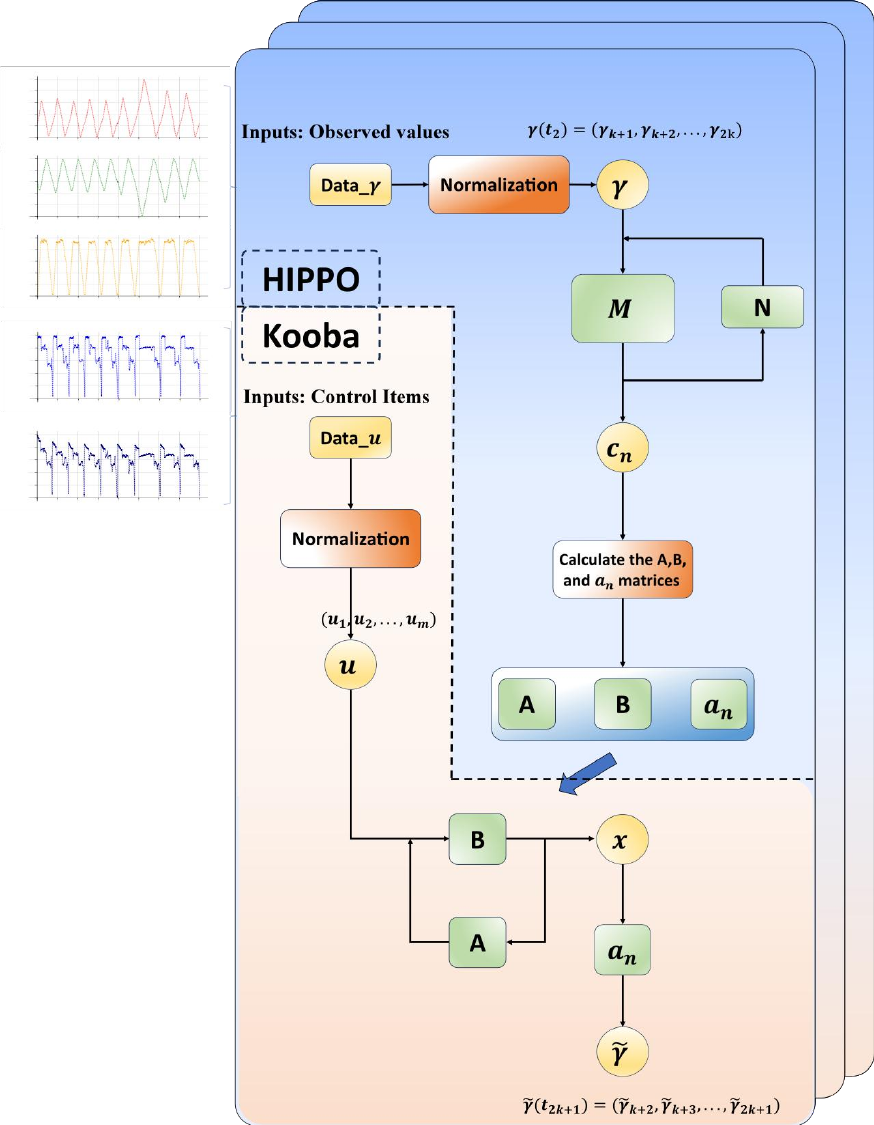}
\caption{The structure of FlightKooba. FlightKooba consists of two components: the first component (HIPPO) computes the coefficients corresponding to the polynomial that fits the system's observation function; the second component (Kooba) generates the corresponding Koopman operator based on the results from the first component, which is used for subsequent prediction tasks. The specific matrices $M$, $N$, $A$, $B$, etc., shown in the figure will be detailed in the subsequent sections.}
\label{fig:a}
\end{figure}

This section introduces FlightKooba, an explainable fast FTP model. Our approach is based on the HIPPO method for compressing continuous signals, Koopman theory for modeling complex systems, and state space equations to connect the two, resulting in FlightKooba. Here, we first provide a framework diagram of FlightKooba, and then explain each part in turn. As shown in Figure \ref{fig:a}

\subsection{The Hippo Theory}

Albert Gu proposed the HiPPO theory \cite{gu2020hippo}, whose main idea is to use various polynomials to infinitely approximate a function in order to compress continuous information. After a series of derivations, the HIPPO algorithm ultimately obtains an iterative formula to calculate the coefficient $c_n$ in front of each polynomial. In this paper, we draw on this theoretical framework and select a Legendre polynomial with good performance as the first part of FlightKooba.

The value of $c_n$ is related to the number of iterations, that is, it is related to time $t$ and changes with time. Let $(c_0(t_k), c_1(t_k),\cdots, c_n(t_k))$ be denoted as $c_{t_k}$ (the value of coefficient $c_{t_k} = (c_0(t_k), c_1(t_k),\cdots, c_n(t_k))$ at time $t_k$), and the final result is:

\begin{equation}
c_{t_{k+1}}=\overline{N}c_{t_k}+\overline{M}\gamma_k
\label{Eq:1}
\end{equation}

In the equation, $t_k$ denotes the $k$th $\Delta t$ time point, where $\Delta t$ is a hyperparameter (for convenience of calculation, if the sequence length is $T$, we set $\Delta t$ to $\frac{1}{T}$). $\gamma_k$ denotes the function value $\gamma(t_k)$ at time $t_k$, i.e., the actual recorded data sample point. The $n$ in $c_n$ denotes the number of terms used in the polynomial fitting of the objective function. It is important to note that $c_n$ represents the coefficient corresponding to the $n$th term when fitting the function, while $c_{t_k}$ denotes the corresponding value of the coefficient vector $c_{t_k}=(c_0(t_k), c_1(t_k),\cdots, c_n(t_k))$ at time $t_k$. Finally, the matrices $\overline{N}$ and $\overline{M}$ are defined as follows:

\begin{equation}
\overline{N}=\left(I-\frac{\Delta t}{2}N\right)^{-1}\left(I+\frac{\Delta t}{2}N\right)
\end{equation}

\begin{equation}
\overline{M}=\Delta\mathrm{t}\left(I-\frac{\Delta t}{2}N\right)^{-1}M
\end{equation}

Among them, the expressions of the matrices $N$ and $M$ are related to the derivation of the HIPPO algorithm. In the original text \cite{gu2020hippo}, the expressions of matrices $N$ and $M$ corresponding to the two methods $LegT$ and $LegS$ are as follows:
\begin{equation}
(LegT)N_{n,k}=-\frac{1}{\omega}\begin{cases}\sqrt{(2n+1)(2k+1)},&k<n\\(-1)^{n-k}\sqrt{(2n+1)(2k+1)},&k\geq n
\end{cases}
\end{equation}

\begin{equation}
(LegT)M_n=\frac{1}{\omega}\sqrt{2(2n+1)}
\end{equation}

\begin{equation}
(LegS)N_{n,k}=-\begin{cases}\sqrt{(2n+1)(2k+1)},&k<n\\n+1,&k=n\\0,&k>n
\end{cases}
\end{equation}

\begin{equation}
(LegS)M_n=\sqrt{2(2n+1)}
\end{equation}

In $LegT$, $\omega$ denotes the length of the historical window of interest (in subsequent experiments, we will assign this value to seq\_len), which is a hyperparameter. The difference between $LegT$ and $LegS$ is that $LegT$ focuses more on information within a historical window length, while $LegS$ focuses more on information across the entire time series length. 

To speed up the calculation, we further derive equation \ref{Eq:1}, which yields:
\begin{equation}
c_{t_{2k+1}}=\overline{N}^{k-1}c_{t_k}+R\overline{M}\gamma_{t_{2k}}
\end{equation}
In the equation, $R=\left(\overline{N}^{k-1},\overline{N}^{k-2},\cdots,\overline{N},I\right)$. $\gamma_{t_{2k}}=(\gamma_{k+1},\gamma_{k+2},\cdots,\gamma_{2k})^T$ denotes the function value of the observation function starting from time $2k$ and moving backward by $k-1$ steps (a vector with $k$ values, which allows both $c_{t_k}$ and $\gamma_{t_{2k}}$ to represent vectors for clarity), that is, a total of $k$ steps are iterated. Detailed derivation of this multistep iteration method is provided in Appendix B.

\subsection{The Koopman Operator}

The Koopman operator is an infinite-dimensional linear operator that can transform complex nonlinear dynamical systems into linear forms. Provides data-driven analytical tools for complex systems that lack explicit mathematical models, such as fluid dynamics, molecular dynamics, and complex networks. Its core idea is to characterize changes in the state of a system through the evolution of an observation function:
\begin{equation}
\mathcal{K}\varphi(x_t)=\varphi(x_{t+k})
\label{Eq:2}
\end{equation}

Here, $\varphi$ denotes the observation function, and $x_t$ denotes the value of state $x$ at time $t$. The Koopman operator acts on $\varphi$ to obtain the observation function of state $x$ at time $t+k$. In the classic Koopman-based method, general time-invariant control of nonlinear systems is considered:
\begin{equation}
x_{k+1}=f(x_k,u_k)
\end{equation}

In the equation, $x_k$ is the state vector of the system at time $k$; $u_k$ is the input vector of the system at time $k$; $f$ is the nonlinear function describing the dynamic behavior of the nonlinear system. According to the Koopman theory \cite{koopman1931}, the above equation can be expressed as:
\begin{equation}
\mathcal{K}\varphi(x_k,u_k)=\varphi\circ f(x_k,u_k)=\varphi(x_{k+1})
\end{equation}

Here, $\varphi$ remains the observation function, and $\circ$ denotes function composition. According to the design of the Koopman operator in \cite{korda2018}, the system modeled by the Koopman operator in formula \ref{Eq:2} can be described as:
\begin{equation}
z_{k+1}=Az_k+Bu_k
\label{Eq:3}
\end{equation}

\begin{equation}
\hat{x}_k=Cz_k
\end{equation}

Here, $z_{k+1}=\varphi(x_{k}) $ represents the representation of the state vector $x$ under the observation function, and the corresponding Koopman operators are divided into $A$ and $B$. That is:
\begin{equation}
\mathcal{K}=\begin{bmatrix}A&B\\ *&*\end{bmatrix}
\end{equation}

where $\mathcal{K}$ is the Koopman operator, and $*$ indicates that this part can be ignored or filled with zero.

\subsection{Methods For Constructing State Space Equations}

A state equation is a mathematical expression that describes the relationship between the state variables of the system. It is widely used in physics, chemistry, engineering (such as thermodynamics, fluid mechanics, control theory), and other fields to describe the laws governing the changes in the macroscopic properties of substances or systems as a function of state variables. In automatic control theory, state equations are usually differential equations based on physical laws (Newton's laws, Kirchhoff's laws).

State space equations are commonly used to describe linear dynamic systems, especially in control theory and systems engineering. They are mathematical expressions that describe how the state of a system changes over time. Take the simplest state equation as an example. The state equation of a spring-mass damping system is shown below:
\begin{equation}
F(t)=m\ddot{x}+c\dot{x}+kx
\end{equation}

In the equation, $m$ is the mass, $k$ is the spring elasticity coefficient, $c$ is the damping coefficient, $x$ is the displacement, $\dot{x}$ is the velocity, and $\ddot{x}$ is the acceleration.

According to this equation, we can obtain:
\begin{equation}
\begin{cases}x=x_1\\\dot{x}=x_2\\\ddot{x}=-\frac{c}{m}x_2-\frac{k}{m}x_1+\frac{1}{m}F(t)
\end{cases}
\end{equation}

To describe the changes in the state vector of the system, we introduce $X$ to represent the state vector, which can be expressed as follows:
\begin{equation}
\begin{cases}X=\binom{x_1}{x_2}\\\dot{X}=\begin{pmatrix}x_2\\-\frac{c}{m}x_2-\frac{k}{m}x_1+\frac{1}{m}F(t)\end{pmatrix}&\end{cases}
\end{equation}

Writing the above equation $X$ in matrix form similar to \ref{Eq:3}, we obtain:
\begin{equation}
\dot{X}=AX+Bu
\label{Eq:matrix}
\end{equation}

Among them:
\begin{align}
\begin{cases}A=\begin{pmatrix}0&1\\-\frac{k}{m}&-\frac{c}{m}\end{pmatrix}\\B=\begin{pmatrix}0\\\frac{1}{m}\end{pmatrix}\\u=F(t)
\end{cases}
\end{align}

The above two sets of equations give the general form of the state space equations for a spring-mass-damping system.

\subsection{The Construction Method Of FlightKooba}

In the previous three sections, we introduced the HIPPO method, the Koopman theory, and the construction of state space equations. Now we need to combine them.

In the HIPPO method, we obtain:
\begin{equation}
\gamma(t)=\sum_{i=0}^nc_ng_n(t)
\label{Eq:4}
\end{equation}

In the equation, $g_n$ is the normalized polynomial of the Legendre polynomial, and $g_n$ satisfies the following relationship with $p_n$ (where $p_n$ is the Legendre polynomial):
\begin{equation}
g_n(t)=\sqrt{\frac{2n+1}{2}}p_n(t)
\label{Eq:5}
\end{equation}

The Legendre polynomials have the following recursive formula:
\begin{equation}
\frac{d}{dt}p_n(t)=np_{n-1}(t)
\label{Eq:6}
\end{equation}

Substituting equations \ref{Eq:5} and \ref{Eq:6} into equation \ref{Eq:4} yields:
\begin{align}
\gamma(t) &= \sqrt{\frac{1}{2}}c_0p_0(t)+\sqrt{\frac{3}{2}}c_1p_1(t)+\cdots+\sqrt{\frac{2n+1}{2}}c_np_n(t) \\
&=\sqrt{\frac{2n+1}{2}}c_np_n(t)+\cdots+\sqrt{\frac{3}{2}}c_{1}\frac{1}{n!}p_n^{\prime}(t)+\sqrt{\frac{1}{2}}c_0\frac{1}{n!}p_n^{(n)}(t)
\end{align}

That is, if we denote the coefficient $\sqrt{\frac{2k+1}{2}}c_k\frac{1}{n(n-1)...(k+1)}$ as $a_{n-k}$, we obtain the following equation:
\begin{equation}
\gamma(t)=a_0p_n(t)+a_1p_n^{\prime}(t)+\cdots+a_{n-1}p_n^{(n-1)}+a_np_n^{(n)}
\label{Eq:7}
\end{equation}

If we regard the Legendre polynomials $p_n$ in the equation as the state of the system described by the equation under the observation function, then we can refer to the ideas in Section 3.3 to obtain the state variables:
\begin{equation}
x=\begin{bmatrix}x_1=p_n\\x_2=p_n^{\prime}\\x_3=p_n^{\prime\prime}\\\vdots\\x_n=p_n^{(n-1)}\end{bmatrix}
\end{equation}

\begin{equation}
\dot{x}=\begin{bmatrix}x_1^{\prime}\\x_2^{\prime}\\x_3^{\prime}\\\vdots\\x_n^{\prime}\end{bmatrix}=\begin{bmatrix}x_2\\x_3\\x_4\\\vdots\\p_n^{(n)}\end{bmatrix}
\label{Eq:8}
\end{equation}

The expression for $p_n^{(n)}$ in the above equation can be derived from formula \ref{Eq:7}:
\begin{equation}
p_n^{(n)}=\frac{1}{a_n}(\gamma(t)-a_0x_1-a_1x_2-...-a_{n-1}x_n)
\end{equation}

Substituting the above equation into formula \ref{Eq:8} and writing it in matrix form similar to formula \ref{Eq:matrix}, we obtain:
\begin{equation}
\dot{x}=Ax+Bu
\label{Eq:9}
\end{equation}

Among them, matrices $A$ and $B$ are respectively:
\begin{equation}
    A=\begin{bmatrix}0&1&0&...&0\\0&0&1&&0\\\vdots&\vdots&\vdots&\vdots&\vdots\\-\frac{a_0}{a_n}&-\frac{a_1}{a_n}&-\frac{a_2}{a_n}&...&-\frac{a_{n-1}}{a_n}\end{bmatrix}
\end{equation}

\begin{equation}
B=\begin{bmatrix}0\\0\\\vdots\\\frac{1}{a_n}\end{bmatrix}
\end{equation}

where $u$ is the value of $\gamma$ at time $t$. If $u$ consists of multiple terms, then we can obtain: 
\begin{equation}
\gamma(t)=b_0u_0+b_1u_1+\cdotp\cdotp\cdotp+b_mu_m
\end{equation}

The coefficients $(b_0, b_1, \cdots, b_m)$ can be obtained through neural network training. At this point, the $B$ matrix will correspondingly change to:
\begin{equation}
B=\begin{bmatrix}0\\0\\\vdots\\\frac{1}{a_N}\end{bmatrix}[b_0\quad b_1\quad...\quad b_m]
\end{equation}

After obtaining the analytical forms of matrices $A$ and $B$, we can utilize the state transition equation in SSM to compute the state after $k$ steps in actual calculations. Details regarding this process can be found in the appendix. However, there is a problem with equation \ref{Eq:9}: its state variables have $n$ terms, while the right side of formula \ref{Eq:7} has $n+1$ terms. To solve this problem, we can retain the $x_1$ before iteration and append it to the state vector after iteration to obtain the complete $n+1$ state variables.

From this, we can obtain the output equation:
\begin{equation}
\tilde{\gamma}=aX
\end{equation}
where $a=(a_0,a_1,...,a_N )$, $X=(x_1,x)^H$.

\begin{figure}[!hbt]
\centering
\includegraphics[width=0.9\linewidth]{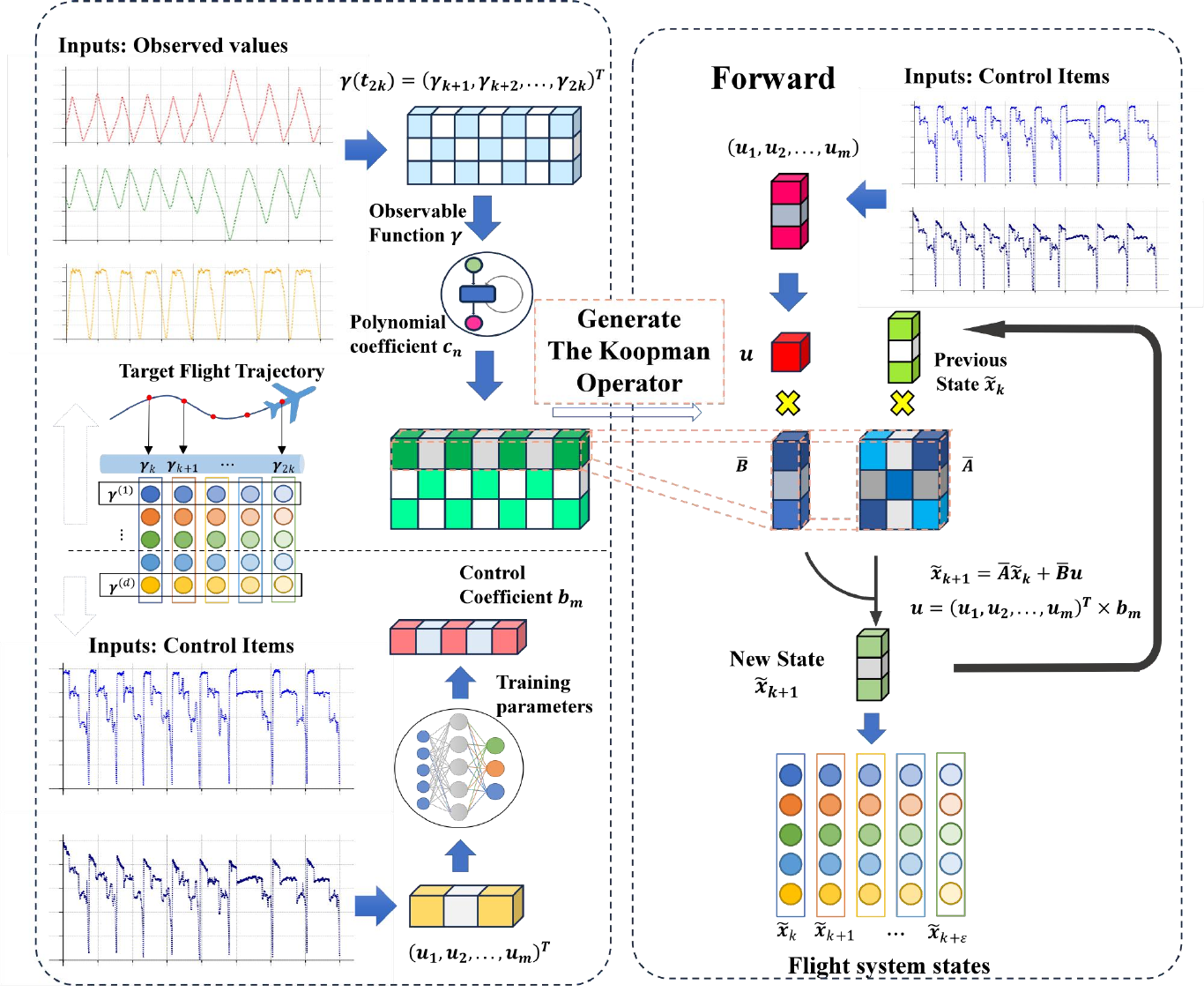}
\caption{Within the framework of flight trajectory prediction applications, FlightKooba computes details. The observation function serves as the state equation, represented by Legendre polynomials. The HIPPO method calculates the coefficients for each polynomial, which are then used to generate the Koopman operator. This operator can predict a sequence of future state changes.}
\label{fig:b}
\end{figure}

Based on the above approach, we can obtain the calculation details of the FlightKooba method. As shown in Figure \ref{fig:b}.

The FlightKooba algorithm is described in the following algorithms \ref{algorithm}.

\begin{algorithm}[!hbt]
\footnotesize
\caption{FlightKooba Algorithm}
\begin{algorithmic}[1]
\Statex \hspace*{-\algorithmicindent} \textbf{Input:} Data sequence $\gamma_k$, Control sequence $u_k$;
\Statex \hspace*{-\algorithmicindent} \textbf{Output:} Prediction sequence $\tilde{\gamma}_{k+\epsilon}$;
\State The data sequence $\gamma_k$ is entered into the HIPPO block and the coefficients $c_n$ are computed by iteratively;
\State Generate the corresponding three matrices $a_n$, $A$, and $B$ based on the calculated coefficient $c_n$;
\State The matrix $\mathcal{K}$ is generated by splicing the generated $A$ and $B$ matrices;
\State The control sequence $u_k$ is fed into FlightKooba and the coefficients $b_m$ are calculated by back propagation;
\State Iterative training is performed to compute the final $b_m$;
\State Calculate the future moment state $\tilde{x}_{k+\epsilon}$ from matrix $\mathcal{K}$ and the input future moment control sequence $u_{k+\epsilon}$;
\State Calculate the value of the future ($\epsilon$ step) by $\tilde{\gamma}_{k+\epsilon} = a_n * \tilde{x}_{k+\epsilon}$ to complete the prediction;
\Statex \textbf{return} Predicted value $\tilde{\gamma}_{k+\epsilon}$
\end{algorithmic}
\label{algorithm}
\end{algorithm}

\section{Data Overview}

\subsection{Flight Trajectory Dataset: CAFUC}

The data set used in this work, named the Multidimensional Flight Trajectory Dataset of the Civil Aviation Flight University of China (CAFUC), from the real world, was generated from realistic fixed-wing aircraft flight training at CAFUC and contains 64 parameters, four artificial labels, 150,000 total frames, 41.6 h of total flight duration in CSV format, 14,356 basic maneuvers and 168 flight training subjects. However, it contains some features with text values or no change; after removing these features, 12 features are obtained that can be used for training. The CAFUC data set is available in the following repository: https://github.com/CAFUC-JJJ/FlightKoopman.

We have already provided a detailed introduction to this data set in our previous article \cite{lu2025flightkoopman}, so we will not go into detail here.

\subsection{Lorentz Attractor Dataset}

The Lorenz attractor was discovered by American meteorologist Edward Lorenz in 1963 while solving a set of nonlinear ordinary differential equations. It is a mathematical model describing atmospheric convection and one of the most iconic examples of chaos theory.

The Lorenz attractor is described by the following set of ordinary differential equations:
\begin{equation}
\begin{cases}\frac{dx}{dt}=\sigma(y-x)\\\frac{dy}{dt}=x(\rho-z)-y\\\frac{dz}{dt}=xy-\beta z
\end{cases}
\end{equation}

Where $x$, $y$, and $z$ are system state variables, and $\sigma$, $\rho$, and $\beta$ are three constants. $\sigma$: Prandtl number, typically taken as a value around 10. $\rho$: Rayleigh Number, controls the convection intensity within the control system. 28 is a commonly used value in the Lorenz system. Under this parameter, the system exhibits typical chaotic behavior. $\beta$: Aspect Ratio, typically set to around 8/3, enabling the system to exhibit complex chaotic behavior. (This dataset is referred to as the Lorenz dataset in the subsequent experimental section.)

CAFUC and Lorenz datasets are primarily used for testing because they are related to physical systems. Similarly, to make the experiment more general, we will also test other classic time-series datasets.

\subsection{Other Time Series Datasets}

The following is a description of this data set: (1) Electricity \cite{velasco2022electric} records electricity consumption data for 321 customers from 2012 to 2014. (2) ETT \cite{zhou2021ETT} consists of oil temperature and power load data recorded from power transformers between July 2016 and July 2018.(3) Exchange \cite{abedin2025exchange} contains daily exchange rate data for eight countries from 1990 to 2016. (4) ILI \cite{peng2023ILI} includes weekly reported percentages of influenza-like illness cases from the US Centers for Disease Control and Prevention from 2002 to mid-2021.(5) Traffic \cite{harrou2024traffic} contains hourly road occupancy rate data for highways in the San Francisco Bay Area from January 2015 to December 2016. (6) Weather \cite{zuo2022weather} is data on 21 weather indicators collected every 10 minutes by the Max Planck Institute for Biogeochemistry in 2020.

\section{Experiments}

In this section, we evaluate the performance of the FlightKooba module in forecasting time series. Specifically, we will assess the following under identical hyperparameter settings: (a) the training time and memory consumption of different models across various datasets, and (b) the accuracy demonstrated by different models on these datasets.

For modeling performance evaluation, we compare the modules obtained by our method with two additional modules; furthermore, we compare with two full modeling approaches. Modules: (1) The latest xLSTM model from the LSTM team \cite{xlstm}, (2) The Mamba module based on the State Space Model (SSM) \cite{gu2023mamba}; Full Methods: We used two state-of-the-art modeling approaches for time series or flight trajectory prediction (Koopa and FlightBERT++) \cite{koopa,flightbert++}.

It should be noted that xLSTM is categorized as a module because the official code interface allows users to select the number of blocks within xLSTM, whereas the latter two methods directly provide complete network designs. However, the FlightKoopa module is designed as a single module within a neural network. Therefore, comparisons with the Mamba module and xLSTM module are the primary focus, while experimental results comparing it with the subsequent two complete methods serve only as reference.

\subsection{Overall Results}

For each model: We set the number of modules for each model to 2, which means that each model consists of two stacked blocks. The batch\_size is set to 32 and the seq\_len is set to 8. We use the MSE loss function,the learning rate is set to 0.001, the number of epochs is set to 50, and we will evaluate the performance of each model by taking the average of 10 training runs.

For the data set: We process each data set based on the shape of the time series data (batch\_size, seq\_len, features). The first 70\% of the entire data set is used as a training set, followed by the next 200 data points as a test set. (All models mentioned in this paper were tested on Linux systems equipped with RTX 5880 Ada graphics cards.)

\begin{figure}[!hbt]
    \centering 

    \begin{subfigure}[b]{0.6\linewidth}
        \centering
        \includegraphics[width=\linewidth]{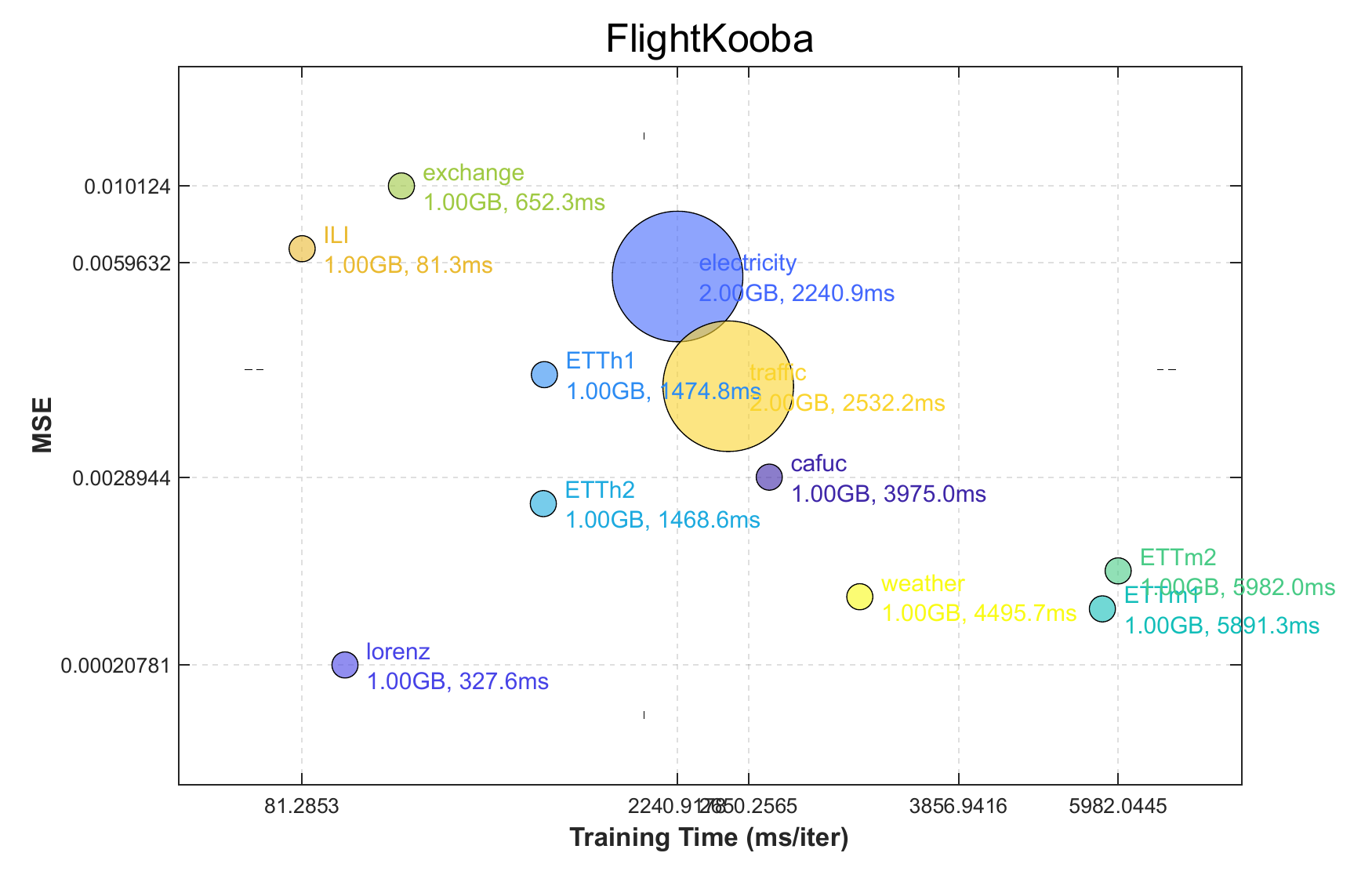}
        \caption{MSE-time-memory bubble plots for FlightKooba across 11 datasets.}
        \label{fig:b1}
    \end{subfigure}
    
    \vspace{1em} 

    \begin{subfigure}[b]{0.9\linewidth}
        \centering
        \includegraphics[width=\linewidth]{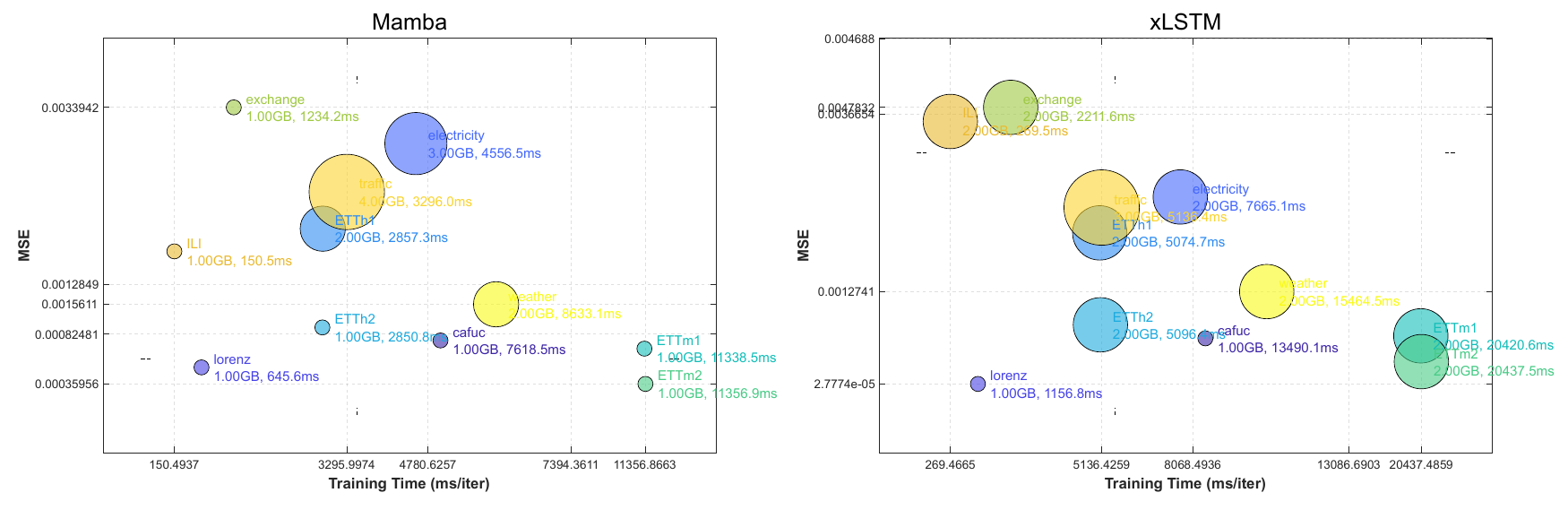}
        \caption{MSE-time-memory bubble plots for Mamba and xLSTM across 11 datasets.}
        \label{fig:b2}
    \end{subfigure}

    \vspace{1em} 

    \begin{subfigure}[b]{0.9\linewidth}
        \centering
        \includegraphics[width=\linewidth]{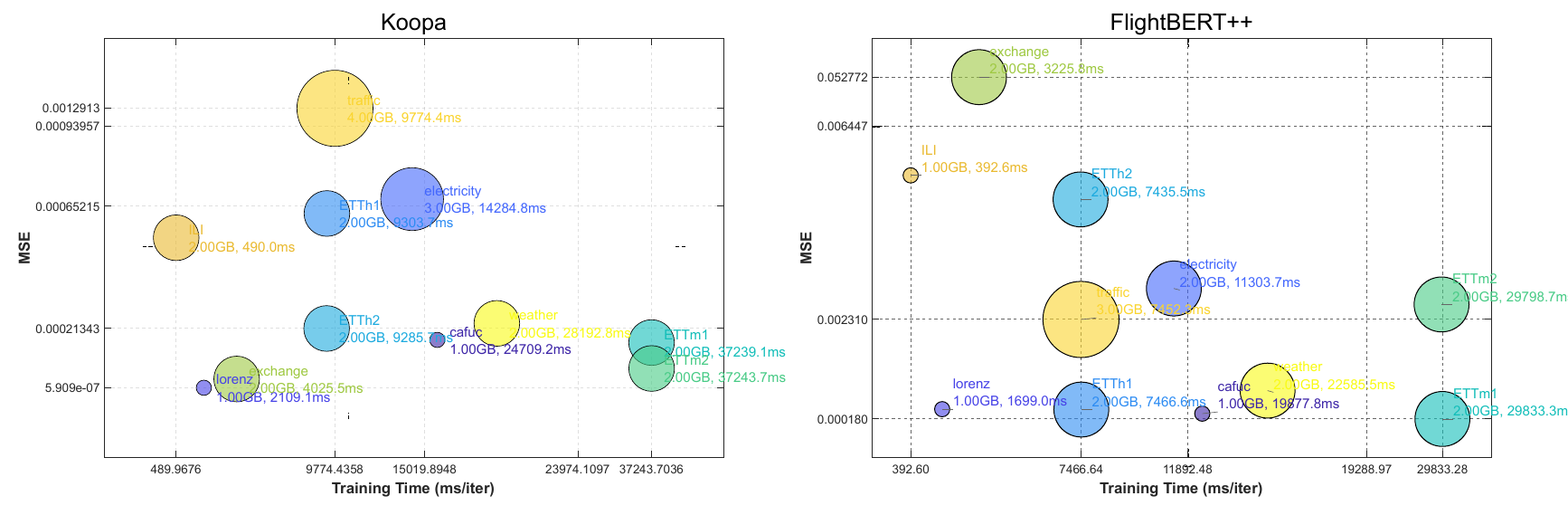}
        \caption{MSE-time-memory bubble plots for Koopa and FlightBERT++ across 11 datasets.}
        \label{fig:b3}
    \end{subfigure}

\end{figure}

In terms of prediction accuracy: Koopa and FlightBERT++ demonstrated the best performance, achieving exceptionally low loss values on the training set. This also highlights the advantages of full neural network modeling approaches in various tasks. However, the training time and memory overhead associated with such modeling methods cannot be overlooked. Regarding training time and memory usage: The FlightKooba module exhibited the shortest training time and the lowest memory consumption among the five models. The experimental results (including MSE, training time, and memory consumption) are shown in Figure \ref{fig:b1}, \ref{fig:b2} and \ref{fig:b3}.

Since the text of the data in the plot is difficult to read, the following table gives all the data in the plot. As shown in Table \ref{tab:1}.

\begin{sidewaystable}[htbp] %
  \centering
  \caption{Module comparison experiment results (training set).}
  \label{tab:1}
  \resizebox{0.95\textheight}{!}{
  \begin{tabular}{
    lccccccccccccccc
  }
    \toprule
    \textbf{Datasets}
      & \multicolumn{3}{c}{\textbf{FlightKooba}}
      & \multicolumn{3}{c}{\textbf{Mamba}}
      & \multicolumn{3}{c}{\textbf{xLSTM}}
      & \multicolumn{3}{c}{\textbf{Koopa}}
      & \multicolumn{3}{c}{\textbf{FlightBERT++}}\\
    \cmidrule(lr){2-4} \cmidrule(lr){5-7} \cmidrule(lr){8-10} \cmidrule(lr){11-13} \cmidrule(lr){14-16}
      & {MSE} & {Times} & {Memory}
      & {MSE} & {Times} & {Memory}
      & {MSE} & {Times} & {Memory}
      & {MSE} & {Times} & {Memory}
      & {MSE} & {Times} & {Memory}\\
    \midrule
    CAFUC      & 0.0029 & {\bfseries 2768.27} & {\bfseries 1.00}
               & 0.0008 & 5004.75           & 1.00
               & 0.0006 & 13490.10          & 1.00
               & {\bfseries 0.0002} & 15755.00          & 1.00
               & 0.0003 & 12481.30          & 1.00\\
    Lorenz     & 0.0002 & {\bfseries 327.61} & {\bfseries 1.00}
               & 0.0005 & 645.60 & 1.00
               & 0.0001 & 1156.78           & 1.00
               & {\bfseries 5.9e-7} & 2109.12           & 1.00
               & 0.0004 & 1698.95           & 1.00\\
    Electricity& 0.0058 & {\bfseries 2240.92} & {\bfseries 2.00}
               & 0.0026 & 4556.50           & 3.00
               & 0.0026 & 7665.13          & 2.00
               & {\bfseries 0.0007} & 14284.80          & 3.00
               & 0.0030 & 11303.70          & 2.00\\
    ETTh1      & 0.0044 & {\bfseries 1474.76} & {\bfseries 1.00}
               & 0.0018 & 2857.34 & 2.00
               & 0.0021 & 5074.68          & 2.00
               & 0.0006 & 9303.71          & 2.00
               & {\bfseries 0.0004} & 7466.64          & 2.00\\
    ETTh2      & 0.0025 & {\bfseries 1468.62} & {\bfseries 1.00}
               & 0.0009 & 2850.76 & 1.00
               & 0.0008 & 5096.36          & 2.00
               & {\bfseries 0.0002} & 9285.73          & 2.00
               & 0.0049 & 7435.49          & 2.00\\
    ETTm1      & 0.0010 & {\bfseries 4684.61} & {\bfseries 1.00}
               & 0.0007 & 8724.79 & 1.00
               & 0.0008 & 15402.40          & 2.00
               & 0.0002 & 28284.90          & 2.00
               & {\bfseries 0.0002} & 22436.80          & 2.00\\
    ETTm2      & 0.0016 & {\bfseries 4775.36 } & {\bfseries 1.00}
               & 0.0004 & 8743.13 & 1.00
               & 0.0003 & 15419.30          & 2.00
               & {\bfseries 7.0e-5} & 28289.50          & 2.00
               & 0.0026 & 22402.2          & 2.00\\
    Exchange   & 0.0071 & {\bfseries 652.31} & {\bfseries 1.00}
               & 0.0029 & 1234.16            & 1.00
               & 0.0038 & 2211.59           & 2.00
               & {\bfseries 3.2e-5} & 4025.54           & 2.00
               & 0.0075 & 3225.84           & 2.00\\
    ILI        & 0.0062 & {\bfseries 81.29} & {\bfseries 1.00}
               & 0.0015 & 150.49  & 1.00
               & 0.0036 & 269.47            & 2.00
               & {\bfseries 0.0005} & 489.97            & 2.00
               & 0.0054 & 392.60            & 1.00\\
    Traffic    & 0.0042 & {\bfseries 2532.24} & {\bfseries 2.00}
               & 0.0021 & 3296.00           & 4.00
               & 0.0024 & 5136.43          & 3.00
               & {\bfseries 0.0010} & 9774.44          & 4.00
               & 0.0023 & 7452.31          & 3.00\\
    Weather    & 0.0012 & {\bfseries 3289.06} & {\bfseries 1.00}
               & 0.0011 & 6019.41 & 2.00
               & 0.0013 & 10446.30          & 2.00
               & {\bfseries 0.0002} & 19238.60          & 2.00
               & 0.0009 & 15189.00          & 2.00\\
    \bottomrule
  \end{tabular}
  }
\end{sidewaystable}

The training time in the table is in $ms$ and the memory is in $GB$. Experimental results demonstrate that FlightKooba exhibits significant advantages in terms of memory and computational overhead across all datasets. The lowest mean squared error (MSE) values are achieved by the Koopa and FlightBERT++ models. This is because these two approaches employ full-fledged models with a large number of trainable parameters, resulting in superior fitting performance. In contrast, the first three methods merely stack two modules together, and therefore, their insufficient fitting capability is expected.

Once again, it is important to reiterate that FlightKooba was designed as a module capable of being inserted into any network layer, with optimization targets focused on minimizing memory and computational overhead. This approach inevitably involves some trade-offs in accuracy. Although MSE performance in the training set is slightly inferior, this does not necessarily indicate that the FlightKooba module is ineffective. In contrast, in subsequent dimension-wise predictions, the curves fitted by the FlightKooba module did not show a significant performance gap. (section \ref{acc})

FlightKooba exhibits minimal memory and time overhead during training because we perform extreme compression of trainable parameters during model construction. The table below \ref{tab:2} displays the number of parameters for all models in different datasets.

\begin{table}[htbp]
  \centering
  \caption{Total number of parameters of the model with different datasets.}
  \label{tab:2}
  \begin{tabular}{lccccc}
    \toprule
    datasets      & FlightKooba                 & Mamba     & xLSTM     &Koopa     &FlightBERT++  \\
    \midrule
    CAFUC         & 4  / {\bfseries 320    }    & 66,892    & 65,172    & 122,040  & 182,353    \\
    Lorenz        & 3  / {\bfseries 38     }    & 65,860    & 64,140    & 113,816  & 178,769    \\
    Electricity   & 62 / {\bfseries 40,404 }    & 106,753   & 105,033   & 439,692  & 320,785    \\
    ETTh1         & 5  / {\bfseries 84     }    & 66,247    & 64,527    & 116,900  & 180,113    \\
    ETTh2         & 5  / {\bfseries 84     }    & 66,247    & 64,527    & 116,900  & 180,113    \\
    ETTm1         & 5  / {\bfseries 84     }    & 66,247    & 64,527    & 116,900  & 180,113    \\
    ETTm2         & 5  / {\bfseries 84     }    & 66,247    & 64,527    & 116,900  & 180,113    \\
    Exchange      & 5  / {\bfseries 126    }    & 66,376    & 64,656    & 117,928  & 180,561    \\
    ILI           & 4  / {\bfseries 120    }    & 66,247    & 64,527    & 116,900  & 180,113    \\
    Traffic       & 52 / {\bfseries 110,160}    & 176,542   & 174,822   & 995,840  & 563,153    \\
    Weather       & 14 / {\bfseries 420    }    & 68,053    & 66,333    & 131,292  & 186,385    \\
    \bottomrule
  \end{tabular}
\end{table}

The data format of the FlightKooba column is (number of control items $u$ / total number of parameters). Since FlightKooba requires support from control items $u$, but there are no control items in the dataset, we input the last k features into the model as control items $u$ for training when using the FlightKooba module. This is also one of the reasons why the $MSE$ value is relatively high in FlightKooba.

\subsection{Accuracy} \label{acc}

\begin{figure}[!hbt]
\centering
\includegraphics[width=1\linewidth]{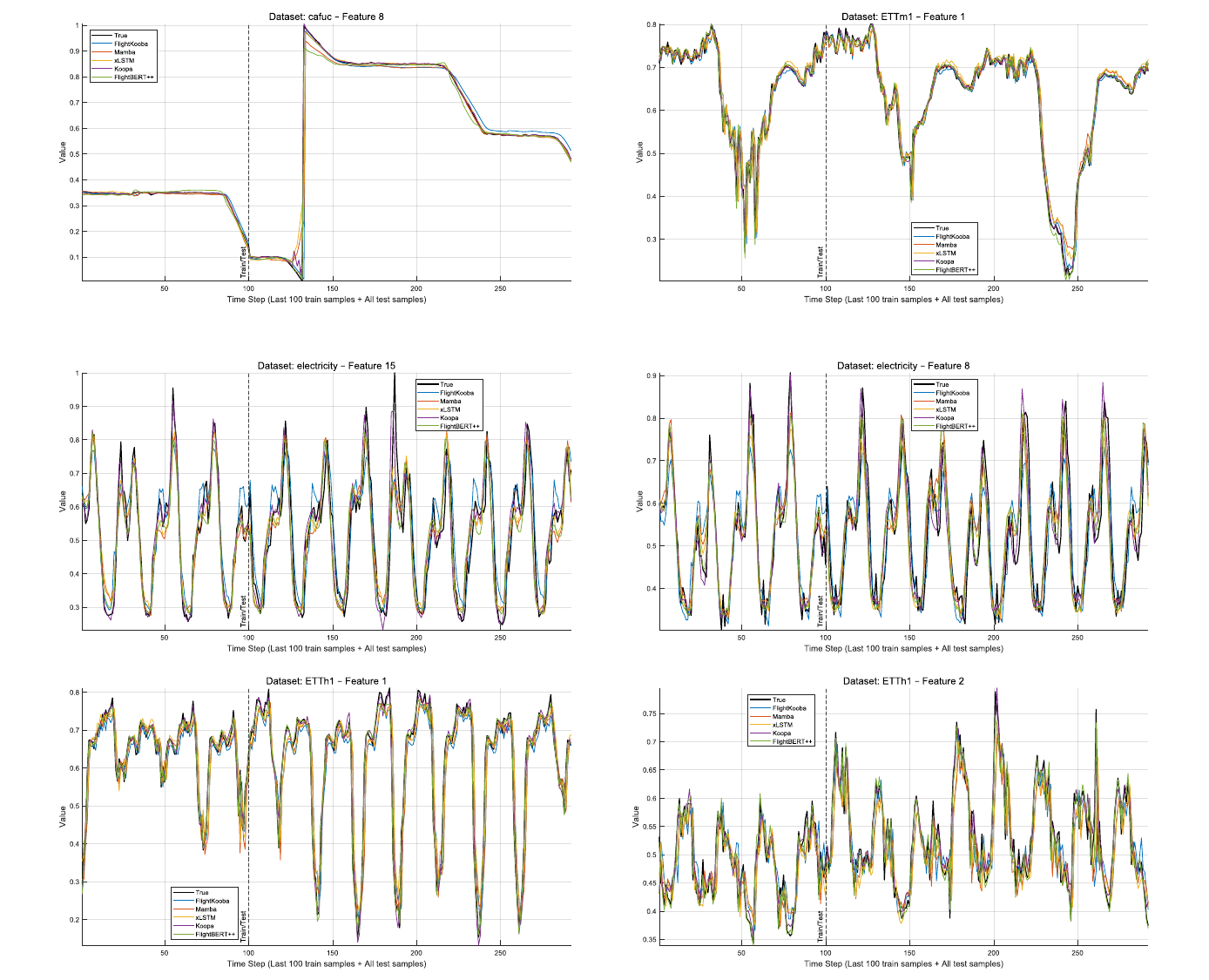}
\caption{Prediction results for some dimensions on the test set.}
\label{fig:f}
\end{figure}

\begin{figure}[!hbt]
\centering
\includegraphics[width=1\linewidth]{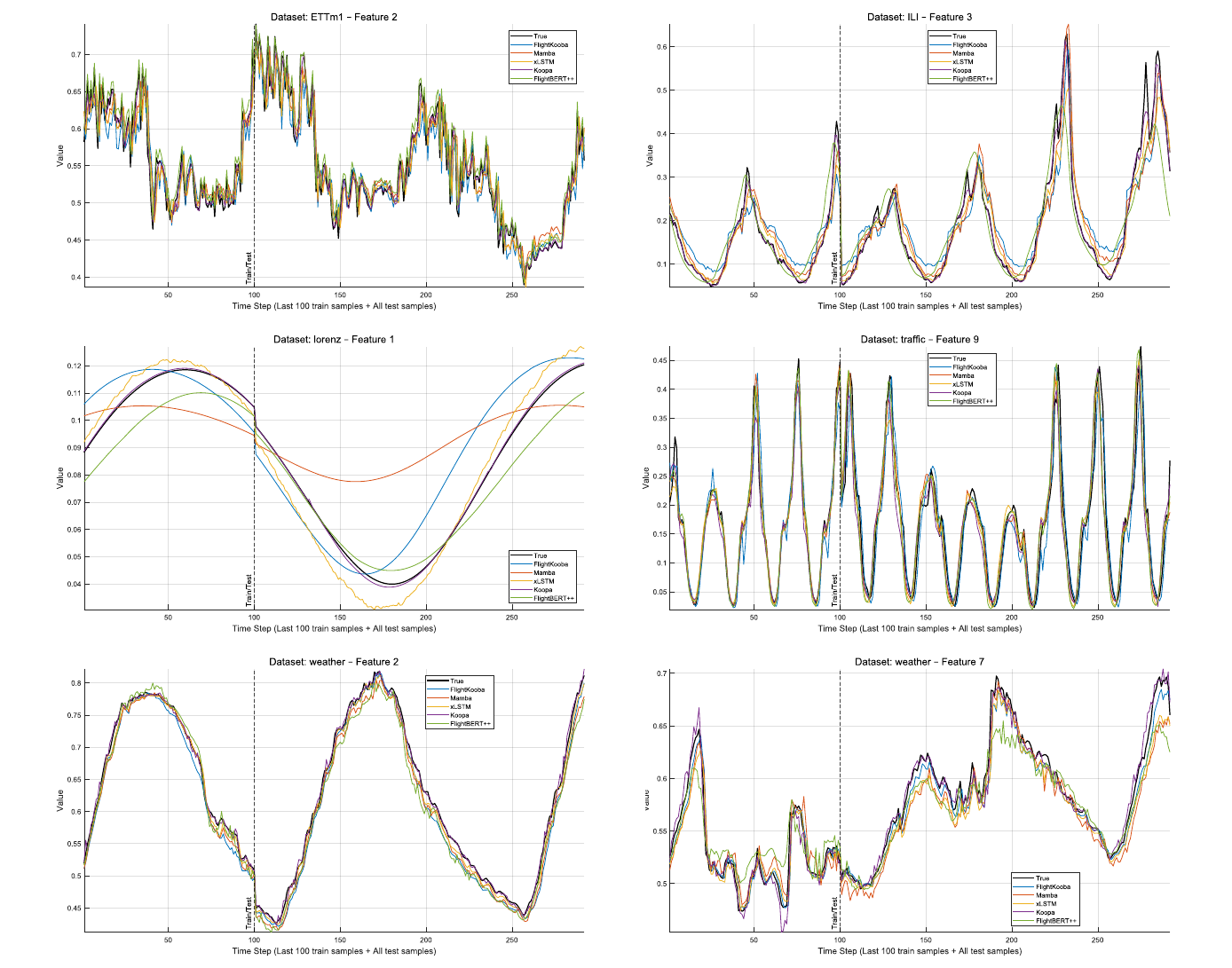}
\caption{Prediction results for some dimensions on the test set.}
\label{fig:g}
\end{figure}

The results of the accuracy test for each model in the test set are shown in Figures \ref{fig:f} and \ref{fig:g}. Due to the large dataset size and feature dimensions, only a subset of results is presented here for demonstration purposes.

The prediction results show that each model performs differently on different datasets, and the same model also performs differently on different features within the same dataset. However, overall, the prediction results of each model are quite good. Among them, the Mamba model performs best in terms of comprehensive performance, while xLSTM and FlightKooba perform well and are not significantly different from each other.

Regarding the ablation experiment: In the FlightKooba structure, regardless of whether the matrix of the HIPPO module or the matrix of the Kooba module is replaced with a trainable parameter matrix, the results are poor, and the model has hardly learned any features of the system. This is because we derived the method for generating Koopman operators from the HIPPO method, which involves the calculation of factorials, making it difficult to learn trainable parameter matrices.

\subsection{Analysis of Model Limitations}

Although comprehensive experiments across 11 datasets (as shown in Tables \ref{tab:1} and \ref{tab:2}) highlight FlightKooba's indisputable computational efficiency—achieving the fastest training speed, lowest memory consumption, and parameter counts several orders of magnitude smaller than advanced baselines like Mamba, xLSTM, and Koopa—our analysis reveals a unique bimodal performance pattern regarding prediction accuracy. Specifically, on test sets, FlightKooba either performs exceptionally well, nearly matching the accuracy of complex models, or fails to capture meaningful dynamics, with few intermediate cases. This section delves into this phenomenon and traces it back to the theoretical foundations of our model.

\begin{figure}[!hbt]
\centering
\includegraphics[width=1\linewidth]{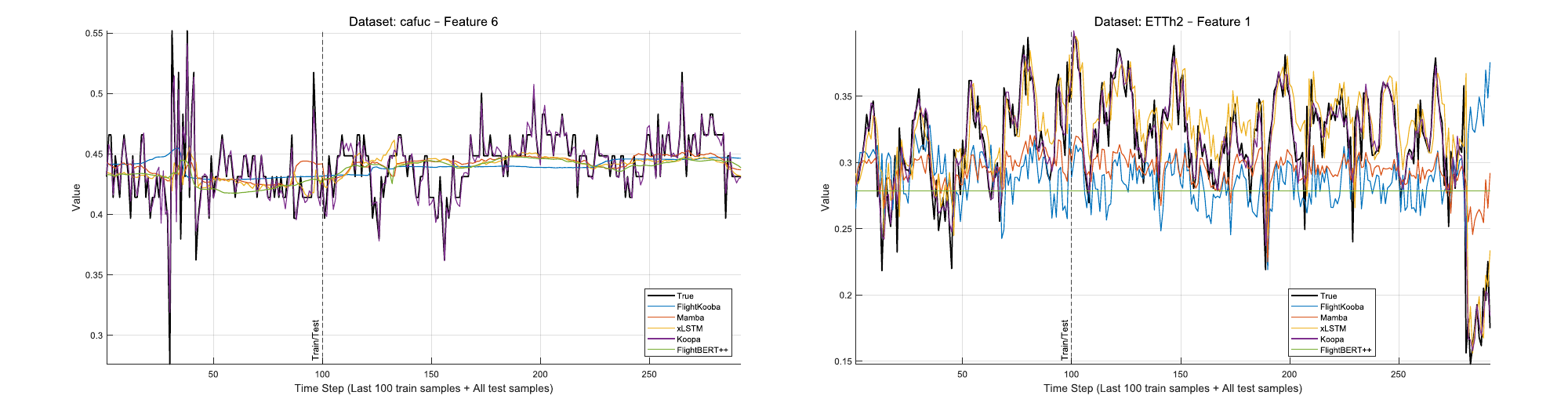}
\caption{Illustration of model limitations on high-frequency datasets. Left: Noise signal from CAFUC (Feature 6), where predictions collapse to the signal mean. Right: Model fails to capture dynamics in ETTh2 (Feature 1), consistent with its low-pass filtering characteristics.}
\label{fig:h2}
\end{figure}

We observe three key behavioral characteristics. First, on datasets dominated by high-frequency components like the sixth feature in CAFUC and ETTh2, FlightKooba's predictive performance sharply deteriorates. The model's output either collapses into a nearly constant line approximating the signal mean or learns only the locations of peaks and troughs with significantly reduced amplitude, indicating its inability to capture underlying dynamics (as shown in \ref{fig:h2}). Second, for periodic signals successfully tracked by the model (e.g., on traffic and electricity datasets), we note persistent phase lags where predicted peaks and troughs trail slightly behind actual values. Third, predicted amplitudes are frequently attenuated, causing forecast curves to appear “flatter” than the true signal.

Initially, these observations cast doubt on the validity of our theoretical derivations. However, after thorough examination and validation, we confirmed that these behaviors are not flaws but direct and predictable consequences of our model's core design. FlightKooba is fundamentally based on the HiPPO framework, which approximates time series using a set of Legendre polynomial bases. From a signal processing perspective, this entire mechanism structurally equates to a powerful low-pass filter. The inherent purpose of a low-pass filter is to suppress high-frequency noise while capturing the primary, smooth long-term trends and periodic components within a signal. This intrinsic property defines both the model's strengths and limitations: its strength lies in effectively filtering out noise to reveal the core smooth skeleton of signal dynamics; its limitation stems from its structural design to inherently ignore high-frequency fluctuations and abrupt transients. This theoretical perspective provides a clear explanation for our experimental results. For signals predominantly composed of high-frequency, noise-like components (such as the sixth feature in CAFUC), the model effectively filters out all dynamic information, leaving only the statistical mean. This also accounts for the observed phase lag and amplitude attenuation—both typical phenomena when applying a low-pass filter to dynamic signals.

Based on this, we can define ideal and unsuitable application scenarios for FlightKooba.
Ideal Application Scenarios:
\begin{itemize}
\item Systems with clear physical underlying structures and fundamental principles (e.g.,  weather system evolution).
\item Signals with strong periodic or seasonal components (e.g., traffic flow, power load, epidemic spread).
\item Applications where computational resources and real-time performance are critical constraints, focusing on core trends rather than fine-grained fluctuations.
\end{itemize}

Unsuitable Application Scenarios:
\begin{itemize}
\item Signals with high-frequency components where random noise itself contains significant information.
\item Chaotic system predictions (e.g., Lorenz systems) where minor perturbations cause substantial long-term divergence.
\item Financial time series (e.g., stock prices) exhibiting behavior closer to random walks.
\end{itemize}

In summary, our findings demonstrate that FlightKooba successfully achieves a highly favorable trade-off between computational cost and prediction performance within its intended application domains, validating its viability as a lightweight and resource-efficient solution. Future work can extend this research in two primary directions. First, exploring FlightKooba as an efficient “backbone extractor” may involve adding a lightweight “high-frequency compensation module” to cover a broader range of signal types. Second, investigating the replacement of the Legendre polynomial basis with alternative functional bases (e.g., wavelets) better suited to specific signal characteristics could enhance the model's flexibility.

\section{Conclusion And Future Works}

In this paper, we introduce FlightKooba, a novel time series modeling approach prioritizing computational efficiency and interpretability. Our comprehensive evaluation across 11 datasets demonstrates its significant breakthrough in this domain. Compared to established modules like Mamba and xLSTM, as well as state-of-the-art models such as Koopa and flightbert++, FlightKooba consistently achieves superior performance in training time, memory usage, and parameter efficiency. This exceptional efficiency stems directly from its fundamental design: FlightKooba avoids extensive parameter training by analytically deriving its state space matrices (A, B, and C) through the synergistic integration of HiPPO and Koopman theory. This approach not only drastically reduces model complexity but also validates the feasibility of systematically constructing Koopman operators from control theory principles, paving a new path for lightweight, interpretable models.

However, as discussed in our analysis, the model's strengths are intrinsically linked to its limitations. We identify three key areas for future research aimed at broadening its applicability:
\begin{itemize}
\item Explicit control variable integration: Due to dataset constraints, our benchmarking utilized proxy control variables (i.e., the last $n$ dimensional features per dataset). A crucial next step is applying FlightKooba to systems with explicit control inputs ($u$), which promises to unlock its full predictive potential and validate its design within a genuine control theory context.
\item Quantification of approximation errors: The model's mathematical derivation relies on approximations inherent to the HiPPO approach and Koopman theory. Systematically quantifying the impact of these approximation errors on final prediction accuracy represents a vital research direction for enhancing model reliability.
\item Enhancing Numerical Stability and Expressiveness: While effective, the use of Legendre polynomials introduces factorial computations that limit model complexity due to numerical precision constraints. Future work should explore alternative polynomial bases or structural optimization modifications to enhance numerical stability and extend the model's ability to capture more complex dynamics without sacrificing efficiency.
\end{itemize}

In summary, we position FlightKooba not as a universal replacement for all time series models, but as a powerful, computationally superior alternative for addressing a specific, critical class of problems. While trade-offs in predictive accuracy exist—particularly for high-frequency signals—its performance in scenarios where efficiency and interpretability are paramount validates its value as a significant contribution to the field.

\bibliographystyle{model1b-num-names}
\bibliography{ref}

\begin{thebibliography}{38}
\expandafter\ifx\csname natexlab\endcsname\relax\def\natexlab#1{#1}\fi
\providecommand{\url}[1]{\texttt{#1}}
\providecommand{\href}[2]{#2}
\providecommand{\path}[1]{#1}
\providecommand{\DOIprefix}{doi:}
\providecommand{\ArXivprefix}{arXiv:}
\providecommand{\URLprefix}{URL: }
\providecommand{\Pubmedprefix}{pmid:}
\providecommand{\doi}[1]{\href{http://dx.doi.org/#1}{\path{#1}}}
\providecommand{\Pubmed}[1]{\href{pmid:#1}{\path{#1}}}
\providecommand{\bibinfo}[2]{#2}
\ifx\xfnm\relax \def\xfnm[#1]{\unskip,\space#1}\fi
\bibitem[{Abedin et~al.(2025)Abedin, Moon, Hassan and Hajek}]{abedin2025exchange}
\bibinfo{author}{M.Z. Abedin}, \bibinfo{author}{M.H. Moon}, \bibinfo{author}{M.K. Hassan}, \bibinfo{author}{P.~Hajek}, \bibinfo{title}{Deep learning-based exchange rate prediction during the covid-19 pandemic}, \bibinfo{journal}{Annals of Operations Research} \bibinfo{volume}{345} (\bibinfo{year}{2025}) \bibinfo{pages}{1335--1386}. \DOIprefix\doi{https://doi.org/10.1007/s10479-021-04420-6}.
\bibitem[{Arbabi and Mezic(2017)}]{arbabi2017hankeldmd}
\bibinfo{author}{H.~Arbabi}, \bibinfo{author}{I.~Mezic}, \bibinfo{title}{Ergodic theory, dynamic mode decomposition, and computation of spectral properties of the koopman operator}, \bibinfo{journal}{SIAM Journal on Applied Dynamical Systems} \bibinfo{volume}{16} (\bibinfo{year}{2017}) \bibinfo{pages}{2096--2126}. \DOIprefix\doi{https://doi.org/10.1137/17M1125236}.
\bibitem[{Beck et~al.(2024)Beck, P{\"o}ppel, Spanring, Auer, Prudnikova, Kopp, Klambauer, Brandstetter and Hochreiter}]{xlstm}
\bibinfo{author}{M.~Beck}, \bibinfo{author}{K.~P{\"o}ppel}, \bibinfo{author}{M.~Spanring}, \bibinfo{author}{A.~Auer}, \bibinfo{author}{O.~Prudnikova}, \bibinfo{author}{M.~Kopp}, \bibinfo{author}{G.~Klambauer}, \bibinfo{author}{J.~Brandstetter}, \bibinfo{author}{S.~Hochreiter}, \bibinfo{title}{xlstm: Extended long short-term memory}, \bibinfo{journal}{arXiv preprint arXiv:2405.04517}  (\bibinfo{year}{2024}). \DOIprefix\doi{https://doi.org/10.48550/arXiv.2405.04517}.
\bibitem[{Budalakoti et~al.(2009)Budalakoti, Srivastava and Otey}]{foqa}
\bibinfo{author}{S.~Budalakoti}, \bibinfo{author}{A.~Srivastava}, \bibinfo{author}{M.~Otey}, \bibinfo{title}{Anomaly detection and diagnosis algorithms for discrete symbol sequences with applications to airline safety}, \bibinfo{journal}{IEEE Transactions on Systems, Man, and Cybernetics, Part C (Applications and Reviews)} \bibinfo{volume}{39} (\bibinfo{year}{2009}) \bibinfo{pages}{101--113}. \DOIprefix\doi{10.1109/TSMCC.2008.2007248}.
\bibitem[{Chen and Lv(2024)}]{chen2024}
\bibinfo{author}{H.~Chen}, \bibinfo{author}{C.~Lv}, \bibinfo{title}{Incorporating eso into deep koopman operator modeling for control of autonomous vehicles}, \bibinfo{journal}{IEEE Transactions on Control Systems Technology}  (\bibinfo{year}{2024}). \DOIprefix\doi{10.1109/TCST.2024.3378456}.
\bibitem[{Gavrilovski et~al.(2016)Gavrilovski, Jimenez, Mavris, Rao, Shin, Hwang and Marais}]{ftp}
\bibinfo{author}{A.~Gavrilovski}, \bibinfo{author}{H.~Jimenez}, \bibinfo{author}{D.N. Mavris}, \bibinfo{author}{A.H. Rao}, \bibinfo{author}{S.~Shin}, \bibinfo{author}{I.~Hwang}, \bibinfo{author}{K.~Marais}, \bibinfo{title}{Challenges and opportunities in flight data mining: A review of the state of the art}, \bibinfo{journal}{AIAA Infotech@ Aerospace}  (\bibinfo{year}{2016}) \bibinfo{pages}{0923}. \DOIprefix\doi{https://doi.org/10.2514/6.2016-0923}.
\bibitem[{Gu and Dao(2023)}]{gu2023mamba}
\bibinfo{author}{A.~Gu}, \bibinfo{author}{T.~Dao}, \bibinfo{title}{Mamba: Linear-time sequence modeling with selective state spaces}, \bibinfo{journal}{arXiv preprint arXiv:2312.00752}  (\bibinfo{year}{2023}). \DOIprefix\doi{https://doi.org/10.48550/arXiv.2312.00752}.
\bibitem[{Gu et~al.(2020)Gu, Dao, Ermon, Rudra and R{\'e}}]{gu2020hippo}
\bibinfo{author}{A.~Gu}, \bibinfo{author}{T.~Dao}, \bibinfo{author}{S.~Ermon}, \bibinfo{author}{A.~Rudra}, \bibinfo{author}{C.~R{\'e}}, \bibinfo{title}{Hippo: Recurrent memory with optimal polynomial projections}, \bibinfo{journal}{Advances in neural information processing systems} \bibinfo{volume}{33} (\bibinfo{year}{2020}) \bibinfo{pages}{1474--1487}. \DOIprefix\doi{https://doi.org/10.48550/arXiv.2008.07669}.
\bibitem[{Gu et~al.(2021)Gu, Goel and R{\'e}}]{gu2021efficiently}
\bibinfo{author}{A.~Gu}, \bibinfo{author}{K.~Goel}, \bibinfo{author}{C.~R{\'e}}, \bibinfo{title}{Efficiently modeling long sequences with structured state spaces}, \bibinfo{journal}{arXiv preprint arXiv:2111.00396}  (\bibinfo{year}{2021}). \DOIprefix\doi{https://doi.org/10.48550/arXiv.2111.00396}.
\bibitem[{Guo et~al.(2024)Guo, Zhang, Yan, Zhang and Lin}]{flightbert++}
\bibinfo{author}{D.~Guo}, \bibinfo{author}{Z.~Zhang}, \bibinfo{author}{Z.~Yan}, \bibinfo{author}{J.~Zhang}, \bibinfo{author}{Y.~Lin}, \bibinfo{title}{Flightbert++: A non-autoregressive multi-horizon flight trajectory prediction framework}, in: \bibinfo{booktitle}{Proceedings of the AAAI Conference on Artificial Intelligence}, volume~\bibinfo{volume}{38}, pp. \bibinfo{pages}{127--134}.
\bibitem[{Ha and Schmidhuber(2018)}]{ha2018}
\bibinfo{author}{D.~Ha}, \bibinfo{author}{J.~Schmidhuber}, \bibinfo{title}{World models}, \bibinfo{journal}{arXiv preprint arXiv:1803.10122}  (\bibinfo{year}{2018}). \DOIprefix\doi{https://doi.org/10.5281/zenodo.1207631}.
\bibitem[{Hafner et~al.(2019)Hafner, Lillicrap, Ba and Norouzi}]{hafner2019}
\bibinfo{author}{D.~Hafner}, \bibinfo{author}{T.~Lillicrap}, \bibinfo{author}{J.~Ba}, \bibinfo{author}{M.~Norouzi}, \bibinfo{title}{Dream to control: Learning behaviors by latent imagination}, \bibinfo{journal}{arXiv preprint arXiv:1912.01603}  (\bibinfo{year}{2019}). \DOIprefix\doi{https://doi.org/10.48550/arXiv.1912.01603}.
\bibitem[{Hafner et~al.(2020)Hafner, Lillicrap, Norouzi and Ba}]{hafner2020}
\bibinfo{author}{D.~Hafner}, \bibinfo{author}{T.~Lillicrap}, \bibinfo{author}{M.~Norouzi}, \bibinfo{author}{J.~Ba}, \bibinfo{title}{Mastering atari with discrete world models}, \bibinfo{journal}{arXiv preprint arXiv:2010.02193}  (\bibinfo{year}{2020}). \DOIprefix\doi{https://doi.org/10.48550/arXiv.2010.02193}.
\bibitem[{Hafner et~al.(2023)Hafner, Pasukonis, Ba and Lillicrap}]{hafner2023}
\bibinfo{author}{D.~Hafner}, \bibinfo{author}{J.~Pasukonis}, \bibinfo{author}{J.~Ba}, \bibinfo{author}{T.~Lillicrap}, \bibinfo{title}{Mastering diverse domains through world models}, \bibinfo{journal}{arXiv preprint arXiv:2301.04104}  (\bibinfo{year}{2023}). \DOIprefix\doi{https://doi.org/10.48550/arXiv.2301.04104}.
\bibitem[{Han et~al.(2022)Han, Euler-Rolle and Katzschmann}]{han2022desko}
\bibinfo{author}{M.~Han}, \bibinfo{author}{J.~Euler-Rolle}, \bibinfo{author}{R.K. Katzschmann}, \bibinfo{title}{Desko: Stability-assured robust control with a deep stochastic koopman operator}, in: \bibinfo{booktitle}{International conference on learning representations (ICLR)}. \DOIprefix\doi{https://doi.org/10.3929/ethz-b-000694283}.
\bibitem[{Harrou et~al.(2024)Harrou, Zeroual, Kadri and Sun}]{harrou2024traffic}
\bibinfo{author}{F.~Harrou}, \bibinfo{author}{A.~Zeroual}, \bibinfo{author}{F.~Kadri}, \bibinfo{author}{Y.~Sun}, \bibinfo{title}{Enhancing road traffic flow prediction with improved deep learning using wavelet transforms}, \bibinfo{journal}{Results in Engineering} \bibinfo{volume}{23} (\bibinfo{year}{2024}) \bibinfo{pages}{102342}. \DOIprefix\doi{https://doi.org/10.1016/j.rineng.2024.102342}.
\bibitem[{Kalman(1960)}]{kalman1960}
\bibinfo{author}{R.E. Kalman}, \bibinfo{title}{A new approach to linear filtering and prediction problems}  (\bibinfo{year}{1960}). \DOIprefix\doi{https://doi.org/10.1115/1.3662552}.
\bibitem[{Koopman(1931)}]{koopman1931}
\bibinfo{author}{B.O. Koopman}, \bibinfo{title}{Hamiltonian systems and transformation in hilbert space}, \bibinfo{journal}{Proceedings of the National Academy of Sciences} \bibinfo{volume}{17} (\bibinfo{year}{1931}) \bibinfo{pages}{315--318}. \DOIprefix\doi{https://doi.org/10.1073/pnas.17.5.315}.
\bibitem[{Korda and Mezi{\'c}(2018)}]{korda2018}
\bibinfo{author}{M.~Korda}, \bibinfo{author}{I.~Mezi{\'c}}, \bibinfo{title}{Linear predictors for nonlinear dynamical systems: Koopman operator meets model predictive control}, \bibinfo{journal}{Automatica} \bibinfo{volume}{93} (\bibinfo{year}{2018}) \bibinfo{pages}{149--160}. \DOIprefix\doi{https://doi.org/10.1016/j.automatica.2018.03.046}.
\bibitem[{Li et~al.(2023)Li, Feng, Yan, Jin, Yang, Sun, Jin and Li}]{li2023}
\bibinfo{author}{F.~Li}, \bibinfo{author}{J.~Feng}, \bibinfo{author}{H.~Yan}, \bibinfo{author}{G.~Jin}, \bibinfo{author}{F.~Yang}, \bibinfo{author}{F.~Sun}, \bibinfo{author}{D.~Jin}, \bibinfo{author}{Y.~Li}, \bibinfo{title}{Dynamic graph convolutional recurrent network for traffic prediction: Benchmark and solution}, \bibinfo{journal}{ACM Transactions on Knowledge Discovery from Data} \bibinfo{volume}{17} (\bibinfo{year}{2023}) \bibinfo{pages}{1--21}. \DOIprefix\doi{https://doi.org/10.1145/3532611}.
\bibitem[{Li et~al.(2019)Li, He, Wu, Katabi and Torralba}]{li2019}
\bibinfo{author}{Y.~Li}, \bibinfo{author}{H.~He}, \bibinfo{author}{J.~Wu}, \bibinfo{author}{D.~Katabi}, \bibinfo{author}{A.~Torralba}, \bibinfo{title}{Learning compositional koopman operators for model-based control}, \bibinfo{journal}{arXiv preprint arXiv:1910.08264}  (\bibinfo{year}{2019}). \DOIprefix\doi{https://doi.org/10.48550/arXiv.1910.08264}.
\bibitem[{Li et~al.(2025)Li, Han and Yin}]{mamko}
\bibinfo{author}{Z.~Li}, \bibinfo{author}{M.~Han}, \bibinfo{author}{X.~Yin}, \bibinfo{title}{Mamko: Mamba-based koopman operator for modeling and predictive control}, in: \bibinfo{booktitle}{The Thirteenth International Conference on Learning Representations}.
\bibitem[{Liu et~al.(2023)Liu, Li, Wang and Long}]{koopa}
\bibinfo{author}{Y.~Liu}, \bibinfo{author}{C.~Li}, \bibinfo{author}{J.~Wang}, \bibinfo{author}{M.~Long}, \bibinfo{title}{Koopa: Learning non-stationary time series dynamics with koopman predictors}, \bibinfo{journal}{Advances in neural information processing systems} \bibinfo{volume}{36} (\bibinfo{year}{2023}) \bibinfo{pages}{12271--12290}.
\bibitem[{Lu et~al.(2025)Lu, Jiang, Bai, Dai and Zhang}]{lu2025flightkoopman}
\bibinfo{author}{J.~Lu}, \bibinfo{author}{J.~Jiang}, \bibinfo{author}{Y.~Bai}, \bibinfo{author}{W.~Dai}, \bibinfo{author}{W.~Zhang}, \bibinfo{title}{Flightkoopman: Deep koopman for multi-dimensional flight trajectory prediction}, \bibinfo{journal}{International Journal of Computational Intelligence and Applications}  (\bibinfo{year}{2025}) \bibinfo{pages}{2450038}. \DOIprefix\doi{https://doi.org/10.1142/S146902682450038X}.
\bibitem[{Morton et~al.(2018)Morton, Jameson, Kochenderfer and Witherden}]{morton2018}
\bibinfo{author}{J.~Morton}, \bibinfo{author}{A.~Jameson}, \bibinfo{author}{M.J. Kochenderfer}, \bibinfo{author}{F.~Witherden}, \bibinfo{title}{Deep dynamical modeling and control of unsteady fluid flows}, \bibinfo{journal}{Advances in Neural Information Processing Systems} \bibinfo{volume}{31} (\bibinfo{year}{2018}). \DOIprefix\doi{https://doi.org/10.48550/arXiv.1805.07472}.
\bibitem[{Peng et~al.(2023)Peng, Ding and Kang}]{peng2023ILI}
\bibinfo{author}{B.~Peng}, \bibinfo{author}{Y.~Ding}, \bibinfo{author}{W.~Kang}, \bibinfo{title}{Metaformer: a transformer that tends to mine metaphorical-level information}, \bibinfo{journal}{Sensors} \bibinfo{volume}{23} (\bibinfo{year}{2023}) \bibinfo{pages}{5093}. \DOIprefix\doi{https://doi.org/10.3390/s23115093}.
\bibitem[{Schmid(2010)}]{schmid2010dmd}
\bibinfo{author}{P.J. Schmid}, \bibinfo{title}{Dynamic mode decomposition of numerical and experimental data}, \bibinfo{journal}{Journal of fluid mechanics} \bibinfo{volume}{656} (\bibinfo{year}{2010}) \bibinfo{pages}{5--28}. \DOIprefix\doi{https://doi.org/10.1017/S0022112010001217}.
\bibitem[{Schulze et~al.(2022)Schulze, Doncevic and Mitsos}]{schulze2022}
\bibinfo{author}{J.C. Schulze}, \bibinfo{author}{D.T. Doncevic}, \bibinfo{author}{A.~Mitsos}, \bibinfo{title}{Identification of mimo wiener-type koopman models for data-driven model reduction using deep learning}, \bibinfo{journal}{Computers \& Chemical Engineering} \bibinfo{volume}{161} (\bibinfo{year}{2022}) \bibinfo{pages}{107781}. \DOIprefix\doi{https://doi.org/10.1016/j.compchemeng.2022.107781}.
\bibitem[{Shi and Meng(2022)}]{shi2022}
\bibinfo{author}{H.~Shi}, \bibinfo{author}{M.Q.H. Meng}, \bibinfo{title}{Deep koopman operator with control for nonlinear systems}, \bibinfo{journal}{IEEE Robotics and Automation Letters} \bibinfo{volume}{7} (\bibinfo{year}{2022}) \bibinfo{pages}{7700--7707}. \DOIprefix\doi{10.1109/LRA.2022.3184036}.
\bibitem[{Velasco et~al.(2022)Velasco, Arnejo and Macarat}]{velasco2022electric}
\bibinfo{author}{L.C.P. Velasco}, \bibinfo{author}{K.A.S. Arnejo}, \bibinfo{author}{J.S.S. Macarat}, \bibinfo{title}{Performance analysis of artificial neural network models for hour-ahead electric load forecasting}, \bibinfo{journal}{Procedia Computer Science} \bibinfo{volume}{197} (\bibinfo{year}{2022}) \bibinfo{pages}{16--24}. \DOIprefix\doi{https://doi.org/10.1016/j.procs.2021.12.113}.
\bibitem[{Verma et~al.(2024)Verma, Heinonen and Garg}]{verma2404}
\bibinfo{author}{Y.~Verma}, \bibinfo{author}{M.~Heinonen}, \bibinfo{author}{V.~Garg}, \bibinfo{title}{Climode: Climate and weather forecasting with physics-informed neural odes, 2024}, \bibinfo{journal}{URL https://arxiv. org/abs/2404.10024}  (\bibinfo{year}{2024}). \DOIprefix\doi{https://doi.org/10.48550/arXiv.2404.10024}.
\bibitem[{Wang et~al.(2024)Wang, Wu, Duan, Zhang, Wang, Peng, Zheng, Liang and Wang}]{wang2024}
\bibinfo{author}{K.~Wang}, \bibinfo{author}{H.~Wu}, \bibinfo{author}{Y.~Duan}, \bibinfo{author}{G.~Zhang}, \bibinfo{author}{K.~Wang}, \bibinfo{author}{X.~Peng}, \bibinfo{author}{Y.~Zheng}, \bibinfo{author}{Y.~Liang}, \bibinfo{author}{Y.~Wang}, \bibinfo{title}{Nuwadynamics: Discovering and updating in causal spatio-temporal modeling}, in: \bibinfo{booktitle}{The Twelfth International Conference on Learning Representations}. \URLprefix \url{https://openreview.net/forum?id=sLdVl0q68X}.
\bibitem[{Williams et~al.(2015{\natexlab{a}})Williams, Kevrekidis and Rowley}]{williams2015(edmd)}
\bibinfo{author}{M.O. Williams}, \bibinfo{author}{I.G. Kevrekidis}, \bibinfo{author}{C.W. Rowley}, \bibinfo{title}{A data--driven approximation of the koopman operator: Extending dynamic mode decomposition}, \bibinfo{journal}{Journal of Nonlinear Science} \bibinfo{volume}{25} (\bibinfo{year}{2015}{\natexlab{a}}) \bibinfo{pages}{1307--1346}. \DOIprefix\doi{https://doi.org/10.1007/s00332-015-9258-5}.
\bibitem[{Williams et~al.(2015{\natexlab{b}})Williams, Kevrekidis and Rowley}]{williams2015edmd}
\bibinfo{author}{M.O. Williams}, \bibinfo{author}{I.G. Kevrekidis}, \bibinfo{author}{C.W. Rowley}, \bibinfo{title}{A data--driven approximation of the koopman operator: Extending dynamic mode decomposition}, \bibinfo{journal}{Journal of Nonlinear Science} \bibinfo{volume}{25} (\bibinfo{year}{2015}{\natexlab{b}}) \bibinfo{pages}{1307--1346}. \DOIprefix\doi{https://doi.org/10.1007/s00332-015-9258-5}.
\bibitem[{Zeng et~al.(2022)Zeng, Chu, Xu, Liu and Quan}]{atc}
\bibinfo{author}{W.~Zeng}, \bibinfo{author}{X.~Chu}, \bibinfo{author}{Z.~Xu}, \bibinfo{author}{Y.~Liu}, \bibinfo{author}{Z.~Quan}, \bibinfo{title}{Aircraft 4d trajectory prediction in civil aviation: a review. aerospace 9 (2): 91}, \bibinfo{year}{2022}. \DOIprefix\doi{https://doi.org/10.3390/aerospace9020091}.
\bibitem[{Zhang et~al.(2023)Zhang, Han and Yin}]{zhang2023}
\bibinfo{author}{X.~Zhang}, \bibinfo{author}{M.~Han}, \bibinfo{author}{X.~Yin}, \bibinfo{title}{Reduced-order koopman modeling and predictive control of nonlinear processes}, \bibinfo{journal}{Computers \& Chemical Engineering} \bibinfo{volume}{179} (\bibinfo{year}{2023}) \bibinfo{pages}{108440}. \DOIprefix\doi{https://doi.org/10.1016/j.compchemeng.2023.108440}.
\bibitem[{Zhou et~al.(2021)Zhou, Zhang, Peng, Zhang, Li, Xiong and Zhang}]{zhou2021ETT}
\bibinfo{author}{H.~Zhou}, \bibinfo{author}{S.~Zhang}, \bibinfo{author}{J.~Peng}, \bibinfo{author}{S.~Zhang}, \bibinfo{author}{J.~Li}, \bibinfo{author}{H.~Xiong}, \bibinfo{author}{W.~Zhang}, \bibinfo{title}{Informer: Beyond efficient transformer for long sequence time-series forecasting}, in: \bibinfo{booktitle}{Proceedings of the AAAI conference on artificial intelligence}, volume~\bibinfo{volume}{35}, pp. \bibinfo{pages}{11106--11115}.
\bibitem[{Zuo et~al.(2022)Zuo, Qiu, Jia, Wang and Li}]{zuo2022weather}
\bibinfo{author}{H.M. Zuo}, \bibinfo{author}{J.~Qiu}, \bibinfo{author}{Y.H. Jia}, \bibinfo{author}{Q.~Wang}, \bibinfo{author}{F.F. Li}, \bibinfo{title}{Ten-minute prediction of solar irradiance based on cloud detection and a long short-term memory (lstm) model}, \bibinfo{journal}{Energy Reports} \bibinfo{volume}{8} (\bibinfo{year}{2022}) \bibinfo{pages}{5146--5157}. \DOIprefix\doi{https://doi.org/10.1016/j.egyr.2022.03.182}.

\end{thebibliography}

\appendix
\section{Multi-step Iteration}
In the HIPPO algorithm, we obtain the following formula:
\begin{equation}
c^\prime(t)=Nc(t)+M\gamma(t)
\label{A15}
\end{equation}

To facilitate iterative solution with a computer, equation \ref{A15} is discretized. First, the left side of the equation is handled by approximating $c^\prime(t)$ in continuous time as the slope between two points $c_{t_{k+1}}$ and $c_{t_k}$, i.e.,:
\begin{equation}
\dot{c}(t_k)\approx\frac{c_{t_{k+1}}-c_{t_k}}{\Delta\mathrm{t}}
\end{equation}

Then, we move on to the discretization on the right side of the equation. Since the data we obtain is discrete, we assume that $\gamma (t)$ is a fixed value between $[t_k,t_{k+1}]$, so we can discretize $\gamma(t)$ as $\gamma_k=\gamma (t_k)$. There are two ways to discretize $c(t)$. One assumes that $c_{t_k}=c(t_k)$ in the interval $[t_k,t_{k+1}]$, and the other assumes that $c_{t_{k+1}}=c(t_k)$, i.e., the two values at the endpoints of the interval. If we consider $c_{t_k}=c(t_k)$, we obtain:

\begin{equation}
\frac{c_{t_{k+1}}-c_{t_k}}{\Delta t}=Nc_{t_k}+M\gamma_k
\label{B2}
\end{equation}

After simplifying the equation, we obtain:
\begin{equation}
c_{t_{k+1}}=(I+\Delta\mathrm{t}N)c_{t_k}+\Delta\mathrm{t}M\gamma_k
\label{B3}
\end{equation}

At this point, if we consider $c_{t_{k+1}}=c(t_k) $, we obtain:
\begin{equation}
\frac{c_{t_{k+1}}-c_{t_k}}{\Delta t}=Nc_{t_{k+1}}+M\gamma_k
\end{equation}

After simplifying the equation, we obtain:
\begin{equation}
c_{t_{k+1}}=(I-\Delta\mathrm{t}N)^{-1}(c_{t_k}+\Delta\mathrm{t}M\gamma_k)
\label{B5}
\end{equation}

With the above two equations \ref{B3} and \ref{B5}, a third method for estimating the value of $c(t_k)$ emerges, which assumes that:
\begin{equation}
\frac{1}{2}({c_{t_{k+1}}+c_{t_k}})=c(t_k)
\label{B6}
\end{equation}

This approximate estimation method is more reasonable than either of the previous two methods, i.e., the frequently mentioned bilinear form. Substituting equation \ref{B6} into equation \ref{B2} yields:
\begin{equation}
\frac{c_{t_{k+1}}-c_{t_k}}{\Delta\mathrm{t}}=N\frac{1}{2}(c_{t_k}+c_{t_{k+1}})+M\gamma_k
\end{equation}

After sorting, we get:
\begin{equation}
c_{t_{k+1}}=\left(I-\frac{\Delta t}{2}N\right)^{-1}\left[\left(I+\frac{\Delta t}{2}N\right)c_{t_k}+\Delta tM\gamma_k\right]
\end{equation}

Here, $ \Delta\mathrm{t}$ is a hyperparameter representing the distance between discrete points, also known as step size.

Therefore, equation \ref{A15} is discretized as follows:
\begin{equation}
c_{t_{k+1}}=\overline{N}c_{t_k}+\overline{M}\gamma_k
\label{B9}
\end{equation}

Among them, the expressions of matrices $\overline{N}$ and $\overline{M}$ are as follows:
\begin{equation}
\overline{N}=\left(I-\frac{\Delta t}{2}N\right)^{-1}\left(I+\frac{\Delta t}{2}N\right)
\end{equation}

\begin{equation}
\overline{M}=\Delta\mathrm{t}\left(I-\frac{\Delta t}{2}N\right)^{-1}M
\end{equation}

Next, we will consider how to perform batch training on equation \ref{B9}. Starting from $k=1$ and setting the initial state $ c_{t_1}=0$, we have:
\begin{align}
c_{t_2}&=\overline{M}\gamma_1 \\
c_{t_3}&=\overline{NM}\gamma_1+\overline{M}\gamma_2\\
c_{t_4}&=\overline{NNM}\gamma_1+\overline{NM}\gamma_2+\overline{M}\gamma_3\\
&\ldots\nonumber\\
c_{t_{k+1}}&=\overline{N}^{k-1}\overline{M}\gamma_1+\overline{N}^{k-2}\overline{M}\gamma_2+\ldots+\overline{NM}\gamma_{k-1}+\overline{M}\gamma_k
\end{align}

For convenience of derivation, we write $c_{t_{k+1}}$ in matrix form:
\begin{align}
c_{t_{k+1}}&=\overline{N}^{k-1}\overline{M}\gamma_1+\overline{N}^{k-2}\overline{M}\gamma_2+\ldots+\overline{NM}\gamma_{k-1}+\overline{M}\gamma_k \nonumber\\
&=(\overline{N}^{k-1},\overline{N}^{k-2},\ldots,\overline{N},I)\overline{M}\gamma(t_k)\nonumber\\
&=R\overline{M}\gamma(t_k)
\label{B16}
\end{align}
where $R=(\overline{N}^{k-1},\overline{N}^{k-2},\ldots,\overline{N},I)$,$\overline{M}=\Delta\mathrm{t}\left(I-\frac{\Delta t}{2}N\right)^{-1}M$. $\gamma(t_k)=(\gamma_1,\gamma_2,\ldots,\gamma_k)^T$ denotes the observed values of $\gamma(t)$ from $\gamma_k$ backward to the discrete values $k-1$ to the discrete value $k$th (a total of values $k$, i.e. a total of $k$ iterations.

When continuing training, $ c_{t_{k+1}} $ is a non-zero value, so we need to obtain a more general iterative formula. Continuing with the above approach, we obtain:
\begin{equation}
c_{t_{2k+1}}=\overline{N}^{k-1}\overline{M}\gamma_k+\overline{N}^{k-2}\overline{M}\gamma_{k+1}+\ldots+\overline{NM}\gamma_{2k-1}+\overline{M}\gamma_{2k}
\end{equation}

It can also be written in matrix form similar to \ref{B16}:
\begin{equation}
c_{t_{2k+1}}=\overline{N}^{k-1}c_{t_k}+R\overline{M}\gamma(t_{2k})
\end{equation}

Therefore, during the actual iteration, it is only necessary to calculate the $k-1$th power of the matrix once and store the result, which can then be used until the iteration ends, thereby reducing the time complexity of the algorithm.

\end{document}